\documentclass[sigconf]{acmart}
\usepackage{multirow}
\usepackage{array}

\AtBeginDocument{%
  \providecommand\BibTeX{{%
    \normalfont B\kern-0.5em{\scshape i\kern-0.25em b}\kern-0.8em\TeX}}}

\setcopyright{acmcopyright}
\copyrightyear{2022} 
\acmYear{2022} 
\setcopyright{acmcopyright}
\acmConference[WSDM '22] {Proceedings of the Fifteenth ACM International Conference on Web Search and Data Mining}{February 21--25, 2022}{Tempe, AZ, USA.}
\acmBooktitle{Proceedings of the Fifteenth ACM International Conference on Web Search and Data Mining (WSDM '22), February 21--25, 2022, Tempe, AZ, USA}
\acmPrice{15.00}
\acmDOI{10.1145/3488560.3498423}
\acmISBN{978-1-4503-9132-0/22/02}

\begin{document}

\title{Sentiment Analysis of Fashion Related Posts in Social Media}



\author{Yifei Yuan}
\affiliation{%
  \institution{The Chinese University of Hong Kong}
  \country{Hong Kong SAR}
}
\email{yfyuan@se.cuhk.edu.hk}

\author{Wai Lam}
\affiliation{%
 \institution{The Chinese University of Hong Kong}
 \country{Hong Kong SAR}}
 \email{wlam@se.cuhk.edu.hk}





\renewcommand{\shortauthors}{Trovato and Tobin, et al.}

\begin{abstract}
  The role of social media in fashion industry has been blooming as the years have continued on. In this work, we investigate sentiment analysis for fashion related posts in social media platforms.  There are two main challenges of this task. On the first place, information of different modalities must be jointly considered to make the final predictions. On the second place, some unique fashion related attributes should be taken into account. While most existing works focus on traditional multimodal sentiment analysis, they always fail to exploit the fashion related attributes in this task. We propose a novel framework that jointly leverages the image vision, post text, as well as fashion attribute modality to determine the sentiment category. One characteristic of our model is that it extracts fashion attributes and integrates them with the image vision information for effective representation. Furthermore, it exploits the mutual relationship between the fashion attributes and the post texts via a mutual attention mechanism. Since there is no existing dataset suitable for this task, we prepare a large-scale sentiment analysis dataset of over 12k fashion related social media posts. Extensive experiments are conducted to demonstrate the effectiveness of our model.
\end{abstract}




\maketitle

\section{Introduction}
Fashion products are characterised by high variability in terms of rapid changing customer preferences. Users always express their sentiment towards fashion products via social media platforms such as Instagram, Snapchat, Twitter, etc. Take Instagram as an example, according to Facebook data, 
fashion, as an ever-lasting hotspot, is ranked the 4th among all the Instagram users’ top interests. Moreover, as shown by {Mention}\footnote{https://mention.com/en/reports/instagram/hashtags/\#4}, the fashion hashtag which is included in around 3.5 million Instagram posts, is ranked the 3rd among all the most used hashtags on Instagram. 
\begin{figure}[]
  \centering
  \includegraphics[width=\linewidth]{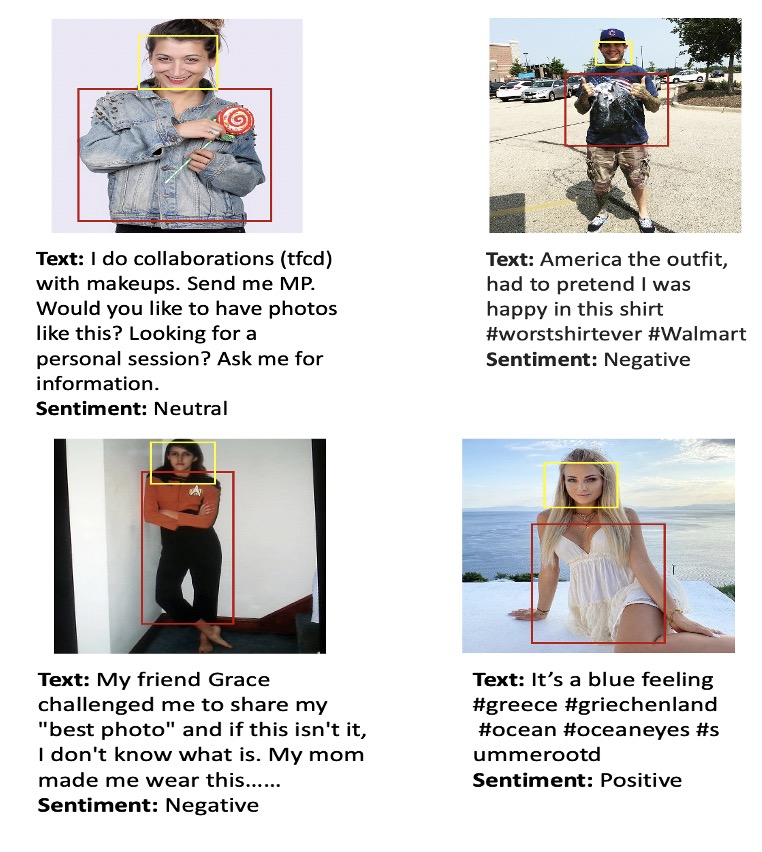}
  \caption{Some examples of our Instagram fashion sentiment analysis dataset. The red box is the fashion items box and the yellow one is the face detection box.}
  \label{dataset}
\end{figure}

The main objective of this paper is to explore social media posts data for recognizing user sentiments towards fashion items. Exploiting the user's sentiment towards fashion items on social media can provide  support for fashion related tasks such as fashion trend understanding and prediction. Besides, it also helps the industry to analyze the fashion trend and thereby enabling fashion brand companies to quickly respond to the ever changing user demands. When viewing a fashion related post, we pay attention to both the texts and images. Information from different modalities must be jointly taken into account to identify the user's real attitude towards the fashion items. For example, as shown in the fashion related social media posts depicted in Figure \ref{dataset}, one must consider both the text and the image to determine the user's sentiment polarity towards the fashion items in the posts. 


However, it is a non-trivial task due to the following challenges: the first one is that the images as well as the texts are equally crucial when determining the user's sentiment towards the fashion items in the posts, which implies that neglecting either of them may give rise to inaccurate prediction. Some typical examples are shown in the first row of Figure \ref{dataset}. Even though the users have smiling faces in both images of the two posts, the first post is apparently a promotion advertisement and does not carry any sentiment. It demonstrates that  despite similar images, different texts can lead to completely different sentiment results. The second row of Figure \ref{dataset} depicts some other examples. Specifically, the text itself may not display much sentiment polarity while some image areas, such as facial expressions, provide evidence for the result. Therefore, how to jointly detect the sentiment polarity in multimodal setting is a difficult but critical task. Another challenge is that, our task is more complex than the typical multimodal sentiment detection task, on which most existing research studies focus. Fashion related posts contain several unique attributes (e.g. style, pattern, color) which play a vital role in sentiment detection. For example, clothes with trendy design and comfy fabric can be easily related to a positive sentiment. Such attributes in fashion related images would provide some clues on the market trends as well as the user preference towards the fashion items. How to make full use of fashion attributes becomes a key instrument in this task.



Although fashion related tasks have always been one of the research hotspots in recent years, previous works mostly focus on tasks such as fashion image analysis including clothing recognition~\cite{liu2016deepfashion,ge2019deepfashion2}, fashion recommendation~\cite{chen2019personalized,jaradat2017deep,hu2015collaborative,kang2017visually}, clothing retrieval~\cite{liao2018interpretable,huang2015cross,hadi2015buy}, etc. Thanks to the development of multimodality techniques, many efforts have been made to link information from different modalities ranging from simple fusion ~\cite{lazaridou2015combining} to more advanced methods such as gated residual connection ~\cite{vo2019composing}, multimodal Transformer ~\cite{tsai2019multimodal,tan2019lxmert,lu2019vilbert}, etc. These techniques have been successfully applied in fashion domain, such as VQA ~\cite{goyal2017making,tapaswi2016movieqa,antol2015vqa},  fashion image retrieval with natural language feedback~\cite{guo2019fashion,10.1145/3404835.3462881}, fashion knowledge extraction ~\cite{ma2019and,ma2019automatic}, etc. However, among all these works, none of them focuses on the fashion multimodal sentiment detection in social media platforms.

In order to tackle this problem, a novel framework is proposed which jointly considers image, post texts, and fashion attributes for predicting the sentiment category. To make full use of the fashion image information, we first extract a bunch of fashion attributes from each image. We reinforce the image vision information by integrating the image vision and fashion attribute features via a fashion-aware vision composition module. We believe that the fashion attributes will help the model have a deeper understanding of the user's sentiment in the posts. 
The second component is denoted as a fashion-aware text composition module. The module makes the prediction based on post texts and fashion attributes where  mutual influences between these two modalities can always be observed in our task. Aiming to reveal such relationship between the representation of texts and the corresponding fashion attribute aspects, a mutual attention mechanism is designed. Furthermore, we propose a vision text composition module to effectively integrate the image vision information with the post text modality, which facilitates better predictions via a multimodal Transformer structure.     

Since there is no existing suitable dataset that captures the user's sentiment polarity towards fashion items in social media posts, we collect and prepare a large-scale dataset based on user-generated social media fashion related posts. We crawl these fashion related posts from a social media platform and annotate the sentiment labels manually. We've published the dataset on Github. 

In conclusion, the main contributions of this work are as follows:
\begin{itemize}
    \item Our model detects the user sentiment polarity from fashion related posts in social media. A novel framework has been developed, which jointly exploits information from images, post texts, and fashion attributes.
    \item The mutual relationship between the fashion attributes extracted from post images and the post texts is captured via a mutual attention mechanism. 
    \item We collect a large-scale dataset of over 12k fashion related social media posts, each includes an image and the corresponding post texts. Extensive experiments are conducted on two datasets to demonstrate the effectiveness of our model.
\end{itemize}

\section{Related Work}
\subsection{Multimodal Sentiment Analysis}
In the past few years, multimodal sentiment analysis has attracted much attention including several tasks such as hate speech detection~\cite{hosseinmardi2015detection}, emotion recognition~\cite{zadeh2018multimodal,jaiswal2020muse}, social media crisis handling~\cite{alam2018crisismmd,agarwal2020crisis}, etc. For the lack of large-scaled dataset, datasets such as MOUD ~\cite{perez2013utterance} MOSI ~\cite{zadeh2016multimodal},  and MOSEI ~\cite{zadeh2018multimodal} are constructed based on product review and recommendation videos from YouTube, each associated with a sentiment label. Following this line, other newly published video based datasets including CH-SIMS~\cite{yu2020ch}, MuSE ~\cite{jaiswal2020muse} allow researchers to further study the interaction between different modalities. In downstream tasks such as multimodal hate speech detection, ~\citet{hosseinmardi2015detection,malmasi2017detecting} aim to look for new approaches to understand and automatically detect incidents of cyberbullying over images in Instagram. On the same problem setting, ~\citet{zhong2016content} collect a dataset  and investigate use of posted images and captions for improved detection of bullying in response to shared content. ~\citet{kiela2020hateful,sabat2019hate} propose a new challenge focused on detecting hate speech in multimodal memes.  Moreover, social media oriented multimodal sentiment analysis tasks such as social media political sentiment prediction~\cite{tumasjan2010predicting}, social media emotion classification~\cite{illendula2019multimodal}, social media target-oriented sentiment analysis~\cite{yu2019adapting} have indicated that information from different modalities can help the model to have a deeper understanding on the classification problem. 

\begin{figure*}
    \centering
    \includegraphics[height=60mm,width=150mm]{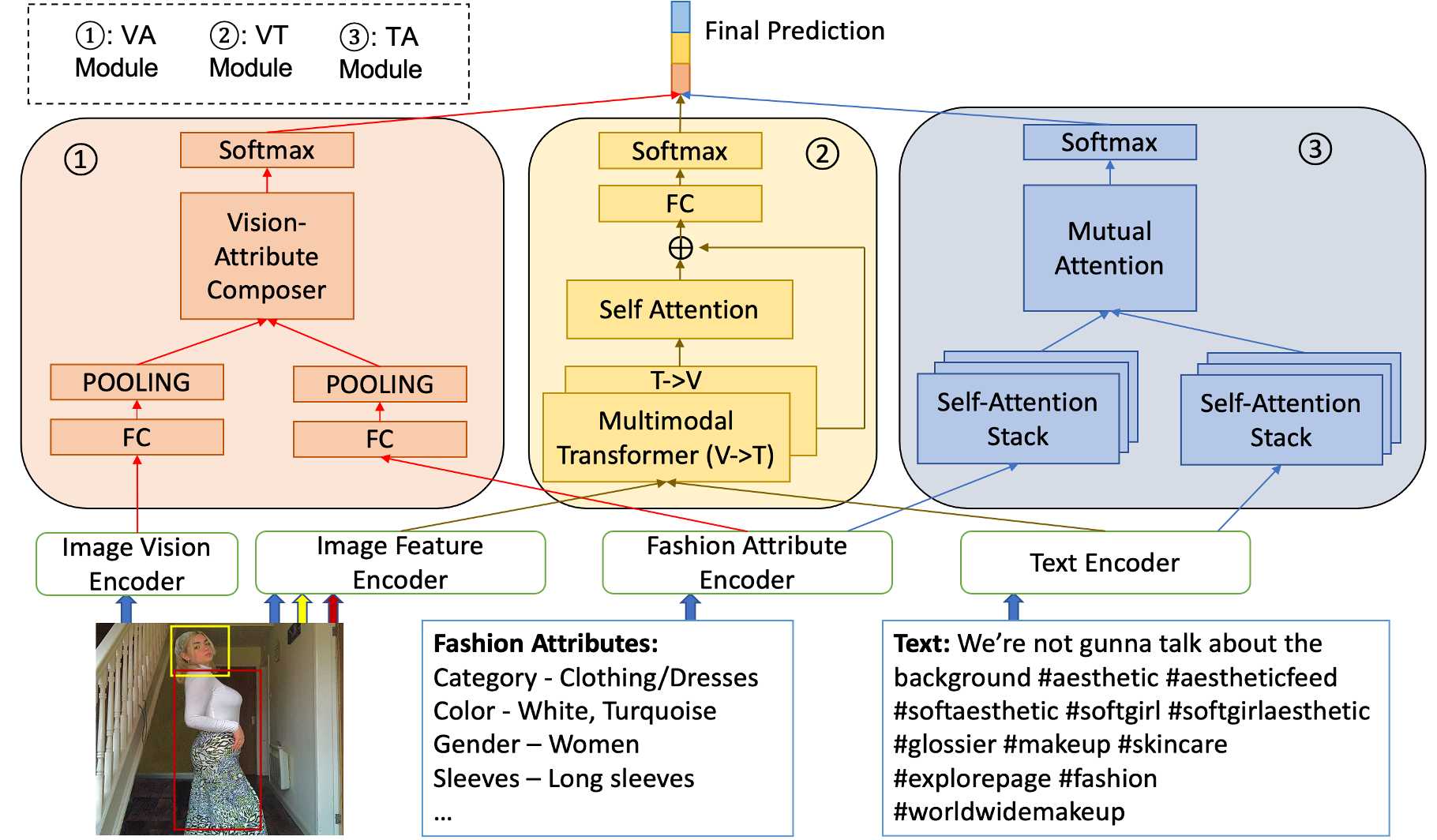}
    \caption{The overall structure of our framework. It consists of a fashion-aware vision composition module (VA module), a fashion-aware text composition module (TA module), and a vision text composition module (VT module). The input is an image annotated with face and fashion item boxes, the corresponding post texts, and the extracted fashion attributes. }
    \label{fig:overall}
\end{figure*}
\subsection{Multimodality in Fashion Related Tasks}
Regarding multimodal fashion related tasks, efforts have been paid in several aspects including outfit recommendation, fashion retrieval, fashion trend prediction, fashion knowledge extraction, etc. In fashion recommendation, for example, ~\citet{li2017mining} score and recommend fashion outfit candidates based on the appearances and metadata, ~\citet{han2017learning} learn a visual-semantic space to not only perform the aforementioned recommendations but also predict the compatibility of a given outfit, ~\citet{chen2019personalized} propose a novel neural architecture for fashion
recommendation based on both image region-level features and user review information, ~\citet{wu2020visual} propose a visual and textual jointly enhanced interpretable model for fashion recommendations. In fashion retrieval, ~\citet{liao2018interpretable} utilize an EI-tree which can cooperate with deep models for end-to-end multimodal learning, ~\citet{gu2018multi,guo2019fashion,10.1145/3404835.3462881} propose a multimodal framework for fashion analysis and data retrieval, ~\citet{ma2020knowledge} forecast the fashion trend by a knowledge enhanced recurrent network model. 

\section{Our Framework}
\subsection{Problem Definition} Our goal is to detect  the sentiment polarity of fashion related social media posts. Each post, including the image and texts, is classified into three categories, namely positive, neutral, and negative. The positive category denotes that the users show strong affection for the fashion items in the post. The negative category refers to those posts where users show a negative attitude towards the fashion items in the post. The neutral category denotes that the user does not show any sentiment polarity towards the fashion items. Some areas in the fashion images which may be related to the user sentiment such as face area, fashion item area are detected in the format of boxes. The training dataset consisting of $N$ data records, is denoted as $\{[img,text,tag]_i\}_{i=1}^{N}$, where $img$ denotes the post image, $txt$ is the post texts, and $tag$ is the sentiment category label.

\subsection{Framework Overview}
Figure \ref{fig:overall} shows the overall structure of our proposed framework. As an initial processing of the images, we extract some fashion attributes associated with each image. The fashion attributes are used to depict the characteristics of the fashion items. Aiming to integrate information from different modalities, our framework is composed of three components, namely fashion-aware vision composition module, fashion-aware text composition module, and vision text composition module. In the fashion-aware vision composition module, we focus on the post images and fashion attributes extracted from them. Specifically, we augment the image information by composing the image vision representation with the fashion attributes. The images and fashion attributes are encoded respectively and then composed together via a multimodal composer. With the fashion attribute information, the model has a deeper understanding on the fashion item characteristics and thus resulting in a more precise prediction.

The fashion-aware text composition module  seeks to make the prediction with the fashion attributes and post texts. Treating the images as the combination of several fashion attributes, we aim to find a correlation between the fashion attributes and the post texts. When dealing with the post texts, some fashion attributes should be more focused. Inversely, some parts in the texts correlated to certain fashion attributes is more important. Take the post text `the pattern is so cute and definitely kills it' and the fashion attribute `floral pattern' as an example, when dealing with the fashion attribute, `the pattern' in the post text is more focused. Inversely, under the same setting, other fashion attributes such as `sleeveless' may not be as important as the pattern attribute. To capture such relationship, a mutual attention mechanism which contains two stages is employed. The  output vector not only represents the post texts under the influence of the fashion attribute aspects, but also integrates the rich information from the other way around.

The vision text composition module aims to integrate image vision information with the corresponding post text features, and learn a dense representation from two different modalities. Following the idea that fashion image areas such as face expression, fashion item style can provide some clues on the user sentiment, we automatically annotate these areas in  the fashion images and encode them together with the post image as image features. The encoded vision and text features are later composed together via a multimodal Transformer followed by a self attention layer. The output is then added by a residual connection contributing to the partial representation which corresponds to the prediction score of the three categories.

\subsection{Preprocessing and Basic Encoding}
\label{Preprocess}
\subsubsection{Image Transformation} Before encoding the images into vectors, we first use some techniques to perform image transformation on post images including random horizontal flip, rotation, translation and scale.
\subsubsection{Image Vision Encoder}
\label{imageencoder}
To encode the post images, we use the ResNet backbone without pretraining on any external data. The image representation encoded by the image vision encoder is denoted as $v_i=ImgEnc(p_i)\in 
\mathbb{R}^{d_p}$, where $d_p$ is the image dimension.
\subsubsection{Image Feature Encoder} Besides the image vision encoder, we also use an image feature encoder which extracts the image features first and then converts them to feature vectors. We first detect the face areas in the images using pretrained MTCNN (Multi-task Cascaded Convolutional Networks) ~\cite{zhang2016joint}. Then we detect the fashion items in the images with YOLOv3 trained with Darknet framework ~\cite{redmon2018yolov3} on DeepFashion2 ~\cite{ge2019deepfashion2} dataset. After that, for each feature area, we use the ResNet backbone same in the image vision encoder. The process that the image feature encoder encodes the $i$-th image can be represented as $v'_i=ImgFtEnc(p_i)\in	\mathbb{R}^{l_v\times{d_p}}$. Where $l_v$ is the number of image features and $d_p$ is the dimension of each image feature.
\subsubsection{Text Preprocessing} 
\label{preprocesstxt}
Before encoding the texts, we preprocess the noisy  user-generated texts. We first convert all the emojis to texts with the python package {demoji}\footnote{https://pypi.org/project/demoji/}. We then remove the external links including the @ and the url links which we assume may not be helpful in our task. Finally, we also detect the text language and translate all the texts into English using {google translation}\footnote{https://pypi.org/project/googletrans/}. 
\subsubsection{Text Encoder} 
After preprocessing, the post texts are tokenized and encoded by a text encoder. The word embeddings are initialized by Glove~\cite{pennington2014glove} pretrained on 6B, 42B, 840B tokens and are concatenated together. After embedded into a vector, the $i$-th text is represented as $e_{i}=TxtEnc(t_i)\in\mathbb{R}^{l_t\times{d_t}}$, where $l_t$ is the number of words in the sentence and $d_t$ is the embedding dimension.
\subsection{Fashion-Aware Vision Composition Module}
\label{VA}
In this module, the first step is the fashion attribute extraction. We extract the fashion attributes from the post images using the fashion tagging tool provided by {Ximilar}\footnote{https://www.ximilar.com/}. We then sort the fashion attributes according to the confidence score and filter out some attributes with the score lower than the preset threshold. The fashion attributes have 20 classes including Category, Top Category, Subcategory, Layers, Style, Cut, Color, Pattern, Age, Material, Length, Neckline, Gender, Sleeves, Gemstones, Fit, Hood, Height, Embellishment, and Type. 

The fashion attributes are in the format of a short text. We encode them using the same embedding strategy as texts, denoted as $f_i=AttrEnc(a_i)\in\mathbb{R}^{l_a\times{d_t}}$,  where $l_a$ is the fashion attribute number (is set to be 20 in our case), and $d_t$ is the word embedding dimension. Note that if one fashion attribute contains more than one words (e.g. color:black and white), the attribute embedding is the average of all the constituent word embeddings.

We then compose the features from image vision and fashion attribute modalities to obtain a joint representation. We choose one of the state-of-the-art mulitimodal composers named ComposeAE~\cite{anwaar2021compositional}, which is initially introduced in the image retrieval task, to learn the compositional representation of image vision and fashion attribute features. In detail, the feature vectors of two modalities are first fed into two pooling layers with fully connected layers and  integrated into a composed form.
\begin{equation}
    c_i=ComposeAE(FC(POOLING(v_i)),FC(POOLING(f_i)))
\end{equation}
where the composed representation $c_i\in\mathbb{R}^d$.

The partial prediction representation is obtained by adding a softmax layer to the representation:
\begin{equation}
V_{final}^{[1]} = softmax(c_i)\in\mathbb{R}^{d_c}
\end{equation}
where $d_c$ is the number of label categories.
\subsection{Fashion-Aware Text Composition Module} 
This module aims to make sentiment predictions based on vision-free information (including fashion attributes and post texts). As mentioned in Section \ref{VA}, the post texts and fashion attributes are encoded by two encoders, forming the text representation $e_i\in\mathbb{R}^{l_t\times{d_t}}$ and the fashion attribute representation $f_i\in\mathbb{R}^{l_a\times{d_t}}$ respectively. In order to obtain information from different granularities, the word embeddings are then fed into a self-attention stack encoder. The encoder is made up of several self-attention blocks which share the same structure. Each block takes the output of the former block as input. Inside the block, each word in the query sentence is attended to words in the key sentence via Scaled Dot-Product Attention. Then the block multiplies the results to the value input and calculates the weighted sum. Finally it adds the vector to the query sentence and feeds it into a fully connected network. We feed the post text and fashion attribute vectors into two self-attention stacks. For a more compact model, the stacks share the same parameters. We then take the average pooling result of the attention blocks as the post text and  fashion attribute encoding results, which can be represented by:
\begin{equation}
    x_i = SAtt(e_i)=Avg(MHAtt(Q,K,V=e_{i,j})_{j=1}^{D})\in\mathbb{R}^{l_t\times{d_t}}
\end{equation}
\begin{equation}
    y_i = SAtt(f_i)=Avg(MHAtt(Q,K,V=f_{i,j})_{j=1}^{D})\in\mathbb{R}^{l_a\times{d_t}}
\end{equation}
where $D$ is the number of attention blocks in the self attention stack. $e_{i,j}$ is the text encoding output of the $j$-th attention block, which also serves as the input of the $j+1$-th block. $MHAtt$ denotes the multi-head attention mechanism. Note that the initial vector $e_{i,0}$ equals to the text embedding $e_i$, and $f_{i,0}$ equals to the attribute embedding result $f_i$.
\subsubsection{Attribute-to-Text Attention}
\label{a2t}
This module contains a mutual attention mechanism which takes the text and fashion attribute output of the self-attention blocks as input and consists of attention from fashion  attribute to text and the other way around. The attribute-to-text attention is employed based on the assumption that each fashion attribute should focus on different parts of the same text sentence. The relatedness between the fashion attributes and the sentence parts are measured by a bunch of weights, which are denoted by the following equations:
\begin{equation}
    \alpha_{mn} = softmax(g(W_1^T[x_{im};y_{in}]+b_1))_n
\end{equation}
where $\alpha_{mn}$ is the attention weight between the $m$-th text feature and the $n$-th fashion attribute feature. $g(\cdot)$ is an non-linear activation function. $softmax(\cdot)_n$ denotes the softmax function is performed along $n$ dimensions. 

With the attentive weights, we multiply them with each text feature and sum them up to generate an attended text representation according to each fashion attribute feature.
\begin{equation}
    x_{i}^{j} = \sum_{k=1}^{l_t}\alpha_{jk}x_{ik}
\end{equation}
where $x_{i}^{j}$ is the attended text vector according to the $j$-th fashion attribute feature. 
The matching vector of the text and the corresponding fashion attribute aspect $a_j$ is defined as:
\begin{equation}
    S(x,a_j) = h(x_{i}^{j},y_{ij})
\end{equation}
where $h(\cdot)$ denotes the inner product, $y_{ij}$ is the $j$-th fashion attribute in the vector.
\subsubsection{Text-to-Attribute Attention}
Following the intuition that for different texts, they should focus on different fashion attributes. We define the weights with respect to the fashion attribute representations and the text representation.
\begin{equation}
    \beta{a_m} = softmax(g(W_2^T[\bar{x_i};y_{im}]+b_2))_m
\end{equation}
where $\beta{a_m}$ is the weight between the $m$-th fashion attribute and the text representation. $\bar{x_i}$ is calculated by averagely pooling all the feature vectors in $x_i$.
The final partial prediction vector is the weighted sum of the correlation vectors obtained in the last step followed by a fully connected layer with a softmax function.
\begin{equation}
    Y_{final}^{[2]} = softmax(FC(\sum_{a_k\in\{a_e,a_s,...\}}\beta{a_k}S(x,a_k)))
\end{equation}
where $S(x,a_k)$ is the attribute-to-text matching  vector obtained in Section \ref{a2t} and $\beta{a_k}$ is the corresponding weight.
\subsection{Vision Text Composition Module} 
In order to integrate information from vision and text  modalities, a multimodal Transformer which is composed of several cross-modal attention blocks, is employed in the module. The Transformer takes the encoded image vision and text features as input, feeds them into the multi-head attention blocks where each block takes the output of the former block as input, and finally outputs the incorporated image vector with text information. Inspired by ~\cite{tsai2019multimodal}, we pass the image vision information to the text language. Each attention block, taking the image vision feature vector $v'_i$, the text feature vector $e_i$ as input, is denoted as: 
\begin{equation}
    X_{t\to{v}} = MHAtt(Q=W_Qv' _i,K=W_Ke_i,V=W_Ve_i)
\end{equation}
where $MHAtt$ denotes multi-head attention, and $W_Q$, $W_K$, $W_V$ are the parameters we intend to learn.
The multimodal Transformer which consists of $L$ attention blocks can be represented as the  follows:
\begin{equation}
\hat{Y}_{t\to{v}}^{[i]} =MHAtt(LN(Y_{t\to{v}}^{[i-1]}),LN(Y_{t}^{[0]}))+LN(Y_{t\to{v}}^{[i-1]})
\end{equation}
\begin{equation}
    Y_{t\to{v}}^{[i]} = f_{\theta}(LN(\hat{Y}_{t\to{v}}^{[i]}))+LN(\hat{Y}_{t\to{v}}^{[i]})
\end{equation}
where $LN$ is the layer normalization function, and $Y_t^{[0]}$ is the text feature. $Y_{t\to{v}}^{[0]}$ is initialized as $X_{t\to{v}}$. $f_\theta$ is a positionwise feed-forward  sublayer parametrized by $\theta$.

The vision and text multimodal representation, which is concatenated by both vision-to-text, and text-to-vision multimodal Transformer vectors, is fed into a self-attention block followed by a residual network. The output is then fed into a fully connected layer with a softmax activation function to form the partial prediction score matrix. This process can be represented by the following equations:
\begin{equation}
    Y_{t,v} = [Y_{t\to{v}}^{[L]};Y_{v\to{t}}^{[L]}]
\end{equation}
\begin{equation}
    Y_{final}^{[3]} = softmax(FC(SAtt(FC(Y_{t,v}))+Y_{t,v}))
\end{equation}
where $SAtt$ denotes the self-attention stack. $FC$ is the fully connected layer. $L$ is the number of attention blocks in the stack. 

\subsection{Partial Prediction Score Combination and Training}
Having the three partial prediction score matrices $Y_{final}^{[1]}$, $Y_{final}^{[2]}$, $Y_{final}^{[3]}$, the final prediction score is the weighted sum of these three matrices, which can be denoted as:
\begin{equation}
    Y_{final} = w_1Y_{final}^{[1]}+w_2Y_{final}^{[2]}+w_3Y_{final}^{[3]}
\end{equation}
where $w_1$, $w_2$, $w_3$ are the weights we would like to optimize. During training, we use the softmax cross entropy as the loss function. 

\section{Experiments}
\subsection{Dataset}
\subsubsection{Our Dataset} Currently, there is no existing dataset for the task of fashion related social media sentiment detection. Social media platforms such as Instagram have seen tremendous growth and  success among the users and brands that have led to the magnification of user-generated content creation and sharing. There are real users who like to share brand experiences and dressing codes, which provide a trustworthy and up-to-date source for fashion sentiment detection. 

We contribute a large-scale fashion related sentiment detection dataset to the community. This dataset contains over 12k fashion related posts from social media, each includes a fashion related image and the  corresponding texts. To obtain the dataset, we collect fashion related posts from Instagram using a set of pre-defined hashtags. Some of these hashtags such as \#fashion, \#ootd mainly contain positive and neutral fashion posts. While some such as \#uglyclothes, \#fashionyoudontunderstand mostly contain fashion posts associated with a negative sentiment. When selecting the hashtags, we first select the top-10 fashion related hashtags from the most popular Instagram fashion hashtags released by best-hashtags.com. We then select hashtags which may contain a negative sentiment from posts of some anti-fashion bloggers. We also collect the posts which do not contain any hashtags or hashtags that are popular but may not have any sentiment polarity such as \#tbt. All the posts were collected in November, 2020.

After collecting the posts, we further clean the data in the following ways. 1) We abandon some posts without any human face and body areas using pre-trained object detection model. 2) We abandon posts with less than 5 tokens or posts only with emojis. 3) If two posts have exact the same image, we remove one of them. 4) For posts with more than 5 hashtags, we only keep the first-5 hashtags and filter out the rest of them. 5) For the noisy texts, we preprocess them using the methods mentioned in Section \ref{preprocesstxt}.

Finally, to obtain ground truth sentiment, we manually conduct annotations and classify all the posts into three categories, namely positive, negative, and neutral. Three annotators are hired to determine the sentiment polarity of all the posts. For each post we make sure that the sentiment label is determined jointly by image and text. After the first-round annotation, we collect the three  annotation results and ask the annotators to discuss and re-annotate the posts whose sentiment results are different. It is worth noting that advertisements or promotions are assumed not to carry any sentiment and are  classified into the neutral category. In our dataset, the average length of the post texts is 25.5 words. Table \ref{table:dataset} shows the number of posts in three categories derived from the top 5 hashtags in our dataset. In agreement with Instagram’s Terms \& Conditions,  we  publish this dataset in the format post IDs. We also publish the preprocess codes and the sentiment label associated with each post.
\begin{table}[h]
\setlength{\abovecaptionskip}{0pt}%
\setlength{\belowcaptionskip}{10pt}%
  \begin{tabular}{ccccc}
    \toprule
    hashtag&Positive&Negative&Neutral&Total\\
    \midrule
    \#fashion & 2489& 739 &1608&4836\\
    \#instagood & 977& 22 &413&1412\\
    \#ootd& 381& 120 &180&681\\
    \#tbt & 187& 89 &111&387\\
    \#stupidshirt & 59& 143 &121&323\\
    \midrule
    \textbf{Total}&4387&2727&4947&12061\\
  \bottomrule
\end{tabular}
\caption{Detailed information of the top 5 hashtags in our dataset.}
\label{table:dataset}
\end{table}
\subsubsection{TWMMS1}In order to further compare the performance of our model with the baselines, we also utilize another dataset named TWMMS1 collected by ~\cite{wang2020cross}. The dataset contains 53701 tweets, each of which consists of an image, the corresponding text, the image attributes, and a target keyphrase considered as a discrete integral label. We borrow the setting from the original paper such that the task is treated as the 4262-labelled classification task and different models are used to predict the target keyphrase label. For tweets that contain more than one keyphrase labels, we only use the first label and abandon the rest of them.
\begin{table*}[t]
\begin{tabular}{p{1.5cm}<{\centering}p{2.5cm}<{\centering}|p{1cm}<{\centering}p{1cm}<{\centering}p{1cm}<{\centering}p{1cm}<{\centering}|p{1.5cm}<{\centering}p{1.5cm}<{\centering}p{1.5cm}<{\centering}p{1.5cm}<{\centering}}
\hline
\multirow{2}{*}{Type}                                                                & \multirow{2}{*}{Model} & \multicolumn{4}{c|}{\textbf{TWMMS1}} & \multicolumn{4}{c}{\textbf{Our Dataset}} \\
                                                                                     &                        & Acc & Precision          & Recall           &F1 &Acc   & Precision     & Recall  &F1       \\ \hline
\multirow{3}{*}{Unimodal}                                                            & Text-Only              &   28.93            &          17.78       &   17.50  & 16.15       &     57.96$\pm$0.66    &     59.92$\pm$1.43 &57.28$\pm$1.20 &56.80$\pm$1.06         \\
                                                  & Image-Only   
                                                       &    15.25           &   8.55              &    7.87       &        8.62 &47.04$\pm$1.11 &46.84$\pm$0.76  &46.38$\pm$0.50 &  44.58$\pm$0.21 \\
                                                                                     & Attribute-Only         &      6.42         & 1.12                &  2.02          &    1.21&43.95$\pm$1.66& 39.75$\pm$0.44&38.49$\pm$0.64 &   34.94$\pm$0.58 \\ \hline
\multirow{6}{*}{\begin{tabular}[c]{@{}l@{}}Multimodal\\ (2 Modalities)\end{tabular}} 
& Early Fusion &31.31&22.66&21.53&19.18&59.68$\pm$1.26&62.41$\pm$1.41&59.96$\pm$1.70&58.91$\pm$1.66\\
    & Late Fusion            &   32.85            &          23.48       &     20.61       &     19.98 & 60.30$\pm$1.06   &      63.09$\pm$2.39        &  60.76$\pm$1.56 &    60.02$\pm$1.63\\
    
& TIRG                   &      29.33         &     18.08            &     18.16       &   18.98       &   62.60$\pm$1.49  &  64.21$\pm$1.06   &  64.95$\pm$1.12        & 63.07$\pm$1.16 \\
                                                        & ComposeAE              &        34.45       & 18.97              & 19.67           &        19.99   &      62.01$\pm$0.51   &
                             64.31$\pm$0.87          &     62.60$\pm$0.85                 &    61.48$\pm$0.71                    \\                                                                      & ViLBERT                &       36.77        &       25.68           &     24.93       &    22.12 & 66.25$\pm$0.63        & 66.06$\pm$0.56    & 66.51$\pm$0.51    &    66.23$\pm$0.19       \\
                                                                                     & ViLBERT CC             &     35.78          &   25.70              &    24.85        &    22.26   &    66.38$\pm$0.99    & 66.13$\pm$0.21 &  67.73$\pm$0.20      &  66.47$\pm$0.41      \\
                            & A2T             &  11.63           &      5.87           &       6.33     &       5.78 &       63.18$\pm$0.83 &       66.16$\pm$1.62 &       63.27$\pm$1.22 &       62.55$\pm$1.07
                        \\
                        & T2A             &    8.36           &        2.86         &      3.42      & 2.90&    62.94$\pm$0.83     &   64.61$\pm$0.20       &     64.89$\pm$0.37         &  63.17$\pm$0.40       \\
                        \hline
\begin{tabular}[c]{@{}l@{}}Multimodal\\ (3 Modalities)\end{tabular}                  & $\rm{M}^{3}$H-Att                &     29.27          &         17.16        &    17.88        &      16.16 &58.08$\pm$2.21&59.85$\pm$2.51&58.36$\pm$2.41&56.86$\pm$2.29        \\ \hline
&Ours                &            37.60            &  26.82             &       26.13          &    24.20        & 67.58$\pm$0.83 & 68.09$\pm$0.76 & 67.13$\pm$0.76   & 67.56$\pm$0.77             \\ \hline
\end{tabular}
\setlength{\abovecaptionskip}{0pt}%
\setlength{\belowcaptionskip}{10pt}%
\caption{Experimental results on two datasets}
\label{tbl:exp}
\end{table*}
\subsection{Experimental Setup}
For evaluation, in both datasets, we split the data into 3 parts: 80\% for training, 10\% for validation, and 10\% for testing. The average results over five runs with different random seed initialization are reported. The evaluation metrics include accuracy, macro recall, macro precision, and macro F1 score. 

With respect to the implementation details, for each image, we resize it into 224*224 dimension. For visual representations, we use the pretrained ResNet-18 as the image encoder backbone, which outputs a 512-D dense vector representation. Regarding text embedding, as mentioned in Section \ref{Preprocess}, the vectors are initialized by Glove, which forms a 900-D text vector representation. We randomly shuffle the order of sentence every epoch for better prediction. We also tried different embedding initialization methods, such as pretrained BERT, RoBERTa, etc, but found no improvements for the final results. In the vision attribute composer, we set the dimension of the output of the fully-connected layer as 2048. In the fashion-aware text composition module, we set the number of attention blocks in the self-attention stack as 3. In the vision text composition module, for the multimodal Transformer, we set the number of multi-head attention blocks as 2 followed by a self-attention layer which only contains one attention block. The confidence threshold in fashion attribute  extraction is set to 0.5.

Our model is implemented with the PyTorch framework. For optimization, we choose the AdamW optimizer with the beta coefficients of (0.55, 0.999). We set the initial learning rate as 0.001. The learning rate drops with a factor of 0.48 on every 10 epochs. By default, the training process is run for 150 epochs with the batch size set to 32. 

\subsection{Compared Methods}
We evaluate several compared methods, which are divided into three classes: unimodal methods, multimodal methods (include 2 modalities), and  multimodal methods (include 3 modalities).
\subsubsection{Unimodal Methods} 
This class includes methods that contain only one modality, including text-only, image-only, and attribute-only methods. For the text-only method, we use BERT as the text encoder and output the prediction with the encoded vector. For the image-only method, we encode the post images with ResNet-152 to make the predictions. For the  attribute-only model, we feed the fashion attribute embedding into a self-attention stack containing 2 attention blocks. The stack serves as the attribute encoder.
\subsubsection{Multimodal Methods (2 Modalities)} This category contains multimodal models which take 2 out of 3 modalities (vision, text, attribute) as input. The first five methods, including Early Fusion, Late Fusion, TIRG ~\cite{vo2019composing}, ComposeAE  ~\cite{anwaar2021compositional}, ViLBERT   ~\cite{lu2019vilbert} are methods composing post image and text features to make the final prediction. ViLBERT CC is the one of the official pretrained versions of ViLBERT which is trained on Conceptual Captions~\cite{lu2019vilbert}. The last two methods take the post text and fashion attribute modalities as input, calculate the attended text-to-attribute or attribute-to-text features, and use the vector with a linear layer to make final prediction. 
\subsubsection{Multimodal Methods (3 Modalities)} We compare our model to a unified framework called Multi-Modality Multi-Head Attention ($\rm{M}^3$H-Att) first proposed by ~\cite{wang2020cross}. This model, which is originally trained on tweet texts and images for social media keyphrase prediction, fuses multimodal features from text, attribute and vision modalities. After feeding the features into some co-attention layers, the output of it is a context vector. We then feed the vector into a sentiment classifier to adapt to our setting. 

\subsection{Experimental Result}
Table \ref{tbl:exp} shows the accuracy, macro precision, macro recall, macro F1 scores of the models mentioned above. As shown in Table \ref{tbl:exp}, on both datasets, our model has better overall performance over all the other models. We observe that for unimodal methods, the text-only classifier outperforms the other two classifiers on both datasets, which implies that the text modality is the most  deterministic in both tasks. The attribute-only method has the worst performance, showing that the fashion attribute does not contain as rich information as the other two modalities.

The multimodal models perform better compared to the unimodal ones. Moreover, it can be observed that the more advanced the fusion strategy is, the better the model performs. For instance, the state-of-the-art methods such as TIRG, ComposeAE have better performance than basic models such as Late Fusion and Early Fusion. In addition, methods containing the Transformer architecture, such as ViLBERT, ViLBERT CC outperform other models when it comes to vision-text fusion. Furthermore, while A2T and T2A attention models do not achieve a better performance in TWMMS1, which may because in the dataset, the text and the attributes do not have a strong relationship. In our dataset, they  have similar and even better performance than most visual-texture models. This demonstrates the mutual attention between fashion attributes and post texts. However, one interesting observation is that, the $\rm{M}^{3}$H-Att model, which takes three modalities as input, does not have a very good performance compared to multimodal models containing 2 modalities in both tasks. One possible reason is that when the three modalities are fed into the model, the model fails to fuse them together in a proper way.
\subsection{Ablation Study}
Table \ref{tbl:ablation} shows the ablation result on the testing set, where VA module represents the fashion-aware vision composition module, TA module represents the fashion-aware text composition module, and VT module represents the vision text composition module. All the numbers are the mathematical average of the results from all  random seeds.
\begin{table}[b]
\setlength{\abovecaptionskip}{0pt}%
\setlength{\belowcaptionskip}{10pt}%
    \centering
    \begin{tabular}{p{2cm}<{\centering}|p{1.2cm}<{\centering}p{1.2cm}<{\centering}p{1.2cm}<{\centering}p{1cm}<{\centering}}
    \hline
        Model & Acc & Precision &Recall &F1  \\
    \hline
         VA module&48.26&47.94&48.18&46.71 \\
         TA module&63.93&65.72&63.69&62.83 \\
         VT module&66.58&66.41&67.96&66.67 \\
    \hline
         w/o VA& 67.09 &67.94&67.15&66.70\\
         w/o TA& 66.07 &66.45&65.92&66.18 \\
         w/o VT& 64.93 & 66.68 &62.98 &64.28\\
    \hline
    \end{tabular}
    \caption{Ablation study of our proposed model}
    \label{tbl:ablation}
\end{table}
\begin{figure*}[]
    \centering
    \includegraphics[width=178mm]{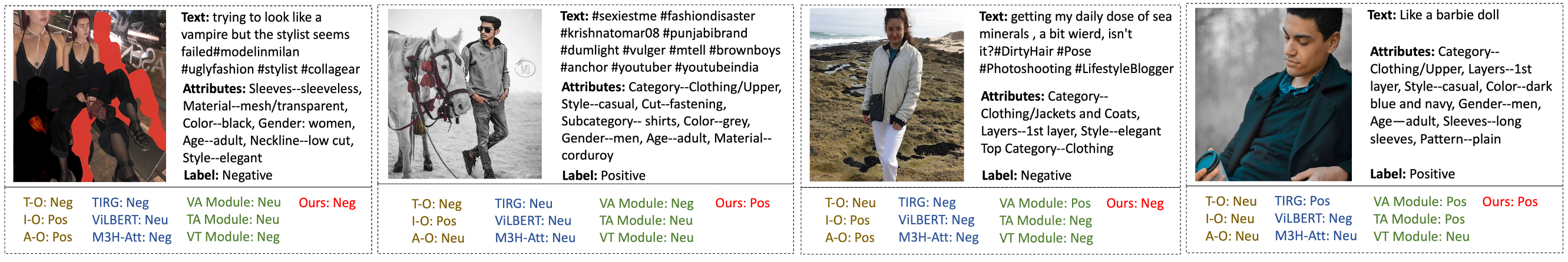}
    \caption{Case study of different methods}
    \label{fig:realcase}
\end{figure*}
As shown in the table, among the three modules, VT module has the best overall performance while VA module has the worst. Together with annotated face and fashion item information, the performance of the VT module is better than the ViLBERT model whose results are recorded in Table \ref{tbl:exp}. The reason that the VA module has the worst performance is that in this module, the post texts are completely abandoned, which contain rich information and can often provide evidence for sentiment prediction. However, in comparison with the image-only method, the performance of the module does improve a lot associated with attribute information (the average testing set F1 score increases from 44.58 in the image-only method to 46.71 in the VA module). This verifies the idea that extracting fashion attributes from the image posts can have a positive effect on the sentiment prediction. 

Considering the performance on models consisting of two modules, we find that for the model without the TA module (which means the model is made up of VA and VT two modules), the performance decreases compared to the single VT module. From this result, we can conclude that the TA module, which contains the mutual attention from the text and fashion attribute, is an indispensable part of the model.  Furthermore, by comparing the results in Table \ref{tbl:ablation} to the final results of our model shown in Table \ref{tbl:exp}, we observe that when the three modules are assembled together, the performance gets improved greatly.

\subsection{Detailed Analysis}
\subsubsection{Performance On Different Sentiment Categories}
\begin{figure}[b]
    \centering
    \includegraphics[width=\linewidth]{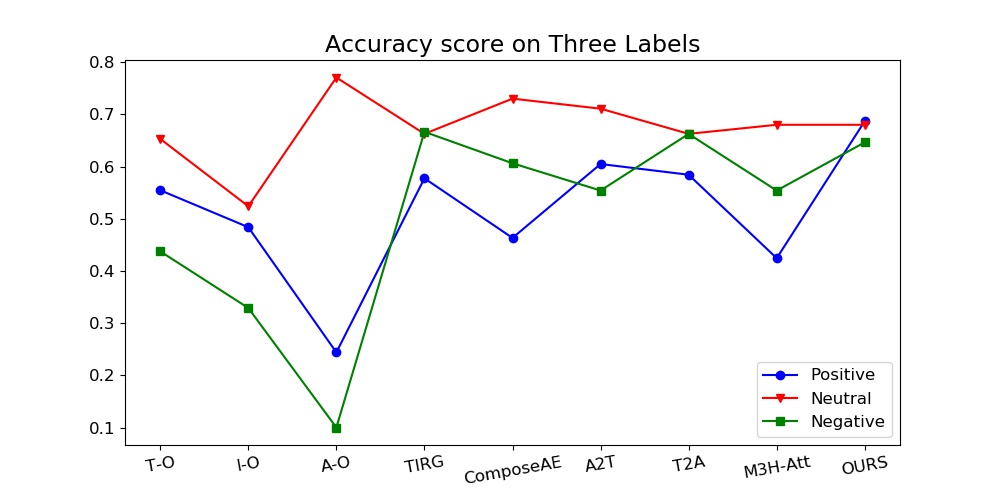}
\caption{Accuracy score of different methods on three sentiment categories. T-O, A-O, I-O denote the three unimodal methods respectively. }
\label{fig:senti}
\end{figure}

Figure \ref{fig:senti} shows the accuracy score of different methods on three sentiment categories. As shown in the figure, on the whole, models have the worst performance on negative posts, which may result from the imbalance of the sample size in the dataset. Specifically, in some methods such as attribute-only, there exists a huge gap between the performance on different categories. It achieves over 0.7 accuracy score in the neutral category while only obtains around 0.1 accuracy score in the negative category. This result shows that fashion items with a neutral sentiment (mostly promotions) may have common fashion attributes and characteristics. Together with information from other modalities, as shown in results of methods such as TIRG and T2A, the performance in the negative category has been greatly improved. It is also worth noting that for some models taking image and text information as input such as TIRG, similar performances are obtained on three categories, with the accuracy score on the neutral samples being around 0.65, on the negative samples being around 0.6, implying that the image vision information plays an important role in the final result prediction. 


Compared to the baseline methods, the performance of our model on three types of data does not differ too much and outperforms most other models. Specifically, the accuracy score of our model in the positive posts is the highest among all the models. Thanks to the three components in the framework, enriched information from three modalities are taken into account together to determine the sentiment category from different perspectives.  


\subsubsection{Real-World Case Study}
In this section, we provide several cases to demonstrate the results. We compare the output of our model with three unimodal methods, three multimodal methods including two multimodal methods with 2 modalities, one multimodal method with 3 modalities. We also list the prediction results of the three components in our model to show the characteristics of different components. 

As shown in Figure \ref{fig:realcase}, for unimodal methods, it tends to be difficult for them to make the right prediction in some cases. For example, in the first case, the post text may be more deterministic in the sentiment prediction since it includes some words such as `uglyfashion', `failed', implying the sentiment polarity. While the post image may be easily associated with a positive sentiment for the reason that the image seems to be taken in a party and the girl in it looks relaxing. In addition, in this  case, it is also hard to make prediction based only on the fashion attributes. The post is misclassified as positive in the attribute-only model since the style attribute `elegant' can often be associated with a happy mood. In the second case, the possible reason that the prediction of the text-only model turning out to be negative is due to the fact that the text contains some negative words including `fashiondisaster'. However, the situation is different in the image-only model. Since in the image, the man is quite enjoying himself, which can be easily associated with a positive sentiment. 

Compared with unimodal methods, multimodal methods take information from different modalities into account, thus leading to a more precise prediction. In the third case, it is easy for multimodal methods to make the right prediction. By contrast, in the fourth case, the outcomes of different methods are not the same. Even though in this case the sentiment category is clear for human annotators, due to the lack of  image areas or text spans that can be directly associated with some sentiment polarities, this kind of posts are more challenging and require models to have a deeper understanding from different modalities. 

\section{Conclusion}
We have explored the task of fashion related post sentiment analysis in social media. The task is constructed based on a multimodal setting such that the sentiment is determined by both post images and associated texts. Different from most existing multimodal models, we extract fashion attributes from the images to help the model better understand the fashion item characteristics. We then propose a novel framework with three components jointly trained to make the final prediction. Since there is no existing dataset, we collect a fashion related sentiment analysis dataset manually labelled with three sentiment labels. Extensive experiments are carried out to demonstrate the effectiveness of our model. 


\bibliographystyle{ACM-Reference-Format}
\bibliography{sample-base}










\end{document}